\documentclass[letterpaper, 10 pt, conference]{ieeeconf}  
\usepackage{url}
\usepackage{cite}
\usepackage{balance}
\usepackage{graphicx}
\usepackage{amsfonts}
\usepackage{fancyhdr}
\usepackage{comment}
\usepackage{times}
\usepackage{amsmath}
\usepackage{changepage}
\usepackage{amssymb}
\usepackage{enumerate}
\usepackage{algorithm}
\usepackage{algorithmic}
\usepackage{bm}
\usepackage{subfigure}
\usepackage{calligra}
\usepackage{multirow}
\usepackage{booktabs}

\IEEEoverridecommandlockouts                              
\title{\LARGE \bf Using DP Towards A Shortest Path Problem-Related Application}

\author{Jianhao Jiao$^{1}$, Rui Fan$^{1}$, Han Ma$^{2}$, Ming Liu$^{1}$
\thanks{This work was supported by the National Natural Science Foundation， under the project No. U1713211, and the Research Grant Council of Hong Kong SAR Government, China, under Project No. 11210017 and No. 21202816, awarded to Prof. Ming Liu. }
\thanks{$^{1}$Jianhao Jiao, Rui Fan and Ming Liu are with the Robotics and Multi-Perception Laborotary, Robotics Institute, The Hong Kong University of Science and Technology, Hong Kong SAR, China.
        {\tt\small \{jjiao, eeruifan, eelium\}}@ust.hk}%
\thanks{$^{2}$Han Ma is with the Department of Precision Instrument, School of Mechanical Engineering, Tsinghua University, Beijing, China. 
	{\tt\small mh15@mails.tsinghua.edu.cn}%
}}

\begin{document}

\maketitle
\thispagestyle{empty}
\pagestyle{empty}

\begin{abstract}
The detection of curved lanes is still challenging for autonomous driving systems. Although current cutting-edge approaches have performed well in real applications, most of them are based on strict model assumptions. Similar to other visual recognition tasks, lane detection can be formulated as a two-dimensional graph searching problem, which can be solved by finding several optimal paths along with line segments and boundaries.
In this paper, we present a directed graph model, in which dynamic programming is used to deal with a specific shortest path problem. This model is particularly suitable to represent objects with long continuous shape structure, e.g., lanes and roads. We apply the designed model and proposed an algorithm for detecting lanes by formulating it as the shortest path problem. To evaluate the performance of our proposed algorithm, we tested five sequences (including 1573 frames) from the KITTI database. The results showed that our method achieves an average successful detection precision of $\textbf{97.5\%}$.
\end{abstract}

\section{INTRODUCTION}
\label{sec.introduction}

\subsection{Motivations}
The major bottleneck in the development of autonomous driving systems is perception problem \cite{thorpe1991toward,liu2012markov}. 
As one of the most essential tasks in perception, traffic scene understanding consists of several computer vision tasks including lane and obstacle detection. 
The functions of lane detection systems are to recognize and localize lane markings on roads as well as provide drivers with a traversable area. 
Consequently, these systems are essential to ensure traffic safety and enhance the driving experience in autonomous vehicles. 
However, lane detection is considered as a demanding task when the actual environment and road conditions such as occlusion, intersections, lights, and weather are considered. 

Several state-of-the-art approaches \cite{ma2018multiple, Lee2017} have achieved excellent performance in real applications. However, most of them are based on precise vanishing point (VP) estimation or a specific road model assumption. 
They do not perform well in challenging environments, such as crossings and turnings, where multiple VPs at different positions exist. 

Inspired by \cite{mortensen1995intelligent,meng2012object,yan2006medical}, similar to these visual recognition tasks, lane detection can also be formulated as a two-dimensional graph searching problem where lane boundaries are extracted through an optimal path search. Hence, the detection of lanes does not need to rely on the strict model assumption. However, finding the best graph model and corresponding shortest path searching algorithm for the given problem is very subtle.

In this paper, we present a directed graph model, in which dynamic programming (DP) is used to solve a specific shortest path problem (SPP). Through a simple pre-processing module, lane boundaries are represented by the designed graph model. We then formulate lane detection as an SPP and solved it effectively by implementing the proposed approach, in which lanes can be localized via the shortest path search. 

\subsection{Contributions}
The contributions of this paper include:
\begin{itemize}
	\item We introduce a graph model where a specific SPP is defined. This graph can intuitively model several objects with a long continuous shape structure e.g., roads.
	\item We analyze the properties of this graph and use DP to solve the well-defined SPP. 
	\item We formulate lane detection as an SPP with this graph model, and then apply the proposed DP-based algorithm to detect multiple lanes in real environments.
\end{itemize}

\subsection{Paper Organization}
The rest of the paper is organized into the following sections: In Section \ref{sec.related_work}, several related work are discussed. In Section \ref{sec.dp_cost_graph}, a general shortest path search problem (SPSP) is presented as well as the corresponding solutions including DP are evaluated using a set of generic test data. Mathematical preliminaries are provided in Section \ref{sec.preliminary}. The application of the proposed algorithm for detecting lanes is introduced in Section \ref{sec.methodology}, followed by the experimental results in Section \ref{sec.experiment}. Finally, Section \ref{sec.conclusion} summarizes the paper and discusses several possible future works.  

\section{RELATED WORK}
\label{sec.related_work}

\subsection{Lane Detection}
The existing lane detection approaches are mainly categorized as feature-based and model-based. The former uses handcraft features, e.g., edges or gradient, to segment lanes boundaries from images, while the latter assumes that the appearance of lanes can be fitted using mathematical models.

The linear \cite{fan_phdthesis}, parabolic \cite{Kreucher1999,zhou2006robust}, and spline models \cite{wang2004lane} are commonly used in representing lanes structure. The linear model performs well in modelling straight lines, while the parabolic and spline models are more flexible when lanes have high curvature. 
However, more parameters introduced into the optimization process will make the computational complexity higher.
Therefore, we use the parabolic model in this paper, which is effective to represent lanes on the inverse perspective mapping (IPM) images.

Produced by the perspective projection, VP is a frequently used model to compute the intersections of lanes on images. 
Traditionally, VP was estimated directly on images by applying Hough transform \cite{fardi2004hough} or texture extraction \cite{kong2009vanishing} techniques.
However, these approaches usually fail in several actual environments. For instance, the accuracy of VP estimation will be greatly reduced on complex roadways. 
Therefore, stereo vision technologies \cite{fan2018real, ozgunalp2014robust} were introduced to address this problem.
Ozgunalp et al. \cite{ozgunalp2014robust} proposed an automatic VP estimation pipeline with a two-step process: firstly, the disparity is estimated from stereo images, and secondly, a corresponding disparity accumulator is computed. Multiple VPs of parallel lanes are found using this accumulator. But different from the above approaches which rely exclusively on VP, our method only uses it to create the IPM images. 

As for the feature-based methods, several approaches proposed in \cite{Wang2008,chapuis2002accurate,zhou2006robust,Bertozzi1998} are based on the edge features. 
Besides, the gradient is another useful feature, which can be used to evaluate the stretch of lanes \cite{fan_phdthesis, cheng2006lane}. Generally, most of the feature-based methods require clear and distinct lane appearance to process, which are sensitive to occlusions and shadow. 
In our method, we adopt the robust Canny operator \cite{canny1986computational} to extract lane boundaries.

Recently, deep learning methods have shown great success in several visual recognition problems \cite{ long2015fully, he2016deep, yun2019fl3d}. Kim et al. \cite{kim2014robust} first explored the performance of convolutional neural network (CNN) for lane detection. Lee et al. \cite{Lee2017} presented a multi-task learning network: VPGNet, for road markings detection, classification, and VP prediction. 
For successfully detecting lanes at a pixel level, several networks for instance segmentation were proposed in \cite{neven2018towards, oliveira2016efficient}. On the other side of the spectrum, the spatial CNN was demonstrated to be accurate in road markings segmentation in \cite{Pan2017}. However, the main limitation is that these methods require a large number of correct manual annotations and much training time.

\subsection{Visual Recognition Based on Shortest Path Problem}
The lane detection problem stated in this paper can be formulated as a graph searching problem, which is degraded as SPP. 
As a general problem, the SPP has been applied in many fields, including path planning \cite{valencia2018path}, computer vision \cite{ulen2015shortest}, and network flow \cite{ahuja2017network}. 
By constructing a graph from raw data, problems can be solved efficiently by applying a general shortest path searching algorithm, such as A* search \cite{hart1968formal}, the Dijkstra's algorithm \cite{dijkstra1959note}, the Floyd-Warshal algorithm \cite{hougardy2010floyd} or the ant colony optimization algorithm \cite{dorigo2009ant}. 
Specifically, SPPs were shown to have great potential in solving visual recognition problems \cite{mortensen1995intelligent, yan2006medical, meng2012object, Zhang2018}. Mortensen et al. \cite{mortensen1995intelligent} presented a DP-based object boundary searching algorithm for image composition, while Meng et al. \cite{meng2012object} employed DP to improve the accuracy of objects segmentation. WIth the stereo vision technologies, Zhang et al. \cite{Zhang2018} proposed a VP constrained model, where the Dijkstra algorithm was implemented to detect road boundaries. However, most of these approaches only focus on the solution of single-pair shortest path problem, which does not apply to our cases.

\section{SHORTEST PATH SEARCHING BASED ON DP}
\label{sec.dp_cost_graph}
In this section, we introduce a specific graph model and show how to use DP to solve an SPP. The results in Section \ref{sec.graph_validation} show that our proposed method can general solve several visual recognition problems, such as the detection of objects with a long continuous shape.

\subsection{Problem Statement} 
We consider a specific directed acyclic graph (DAG) model with a matrix structure. 
As a convention, the designed graph model is denoted by $\mathcal{G}=\big\{\mathcal{P},\mathcal{E}\big\}$, and its dimension is set as $U\times V$.
The nodes are denoted by their positions $(u,v)$, and the directed edges are denoted by their starting/ending node like this: $\big[(s,t),(i,j)\big]$.
In addition, we denote $\mathcal{D}=\big\{(1,V), (2,V),\ldots, (U,V)\big\}$ the set of nodes on the top row. We also denote $\mathcal{F}\big[(i,j), (s,t)\big]$ the cost of $\big[(s,t), (i,j)\big]$. Two important properties of $\mathcal{G}$ are summarized as below:
\begin{itemize}
	\item $\mathcal{G}$ is directed, $\big[(i,j), (s,t)\big]$ and $\big[(s,t), (i,j)\big]$ refer to two different edges.
	\item $\mathcal{G}$ has a bottom-up connection structure, which means that each node has a connection with only $2k+1$ neighbors which are located in its next row.
\end{itemize}

\begin{figure}[h!]
	\centering    
	\includegraphics[width=0.4\textwidth]{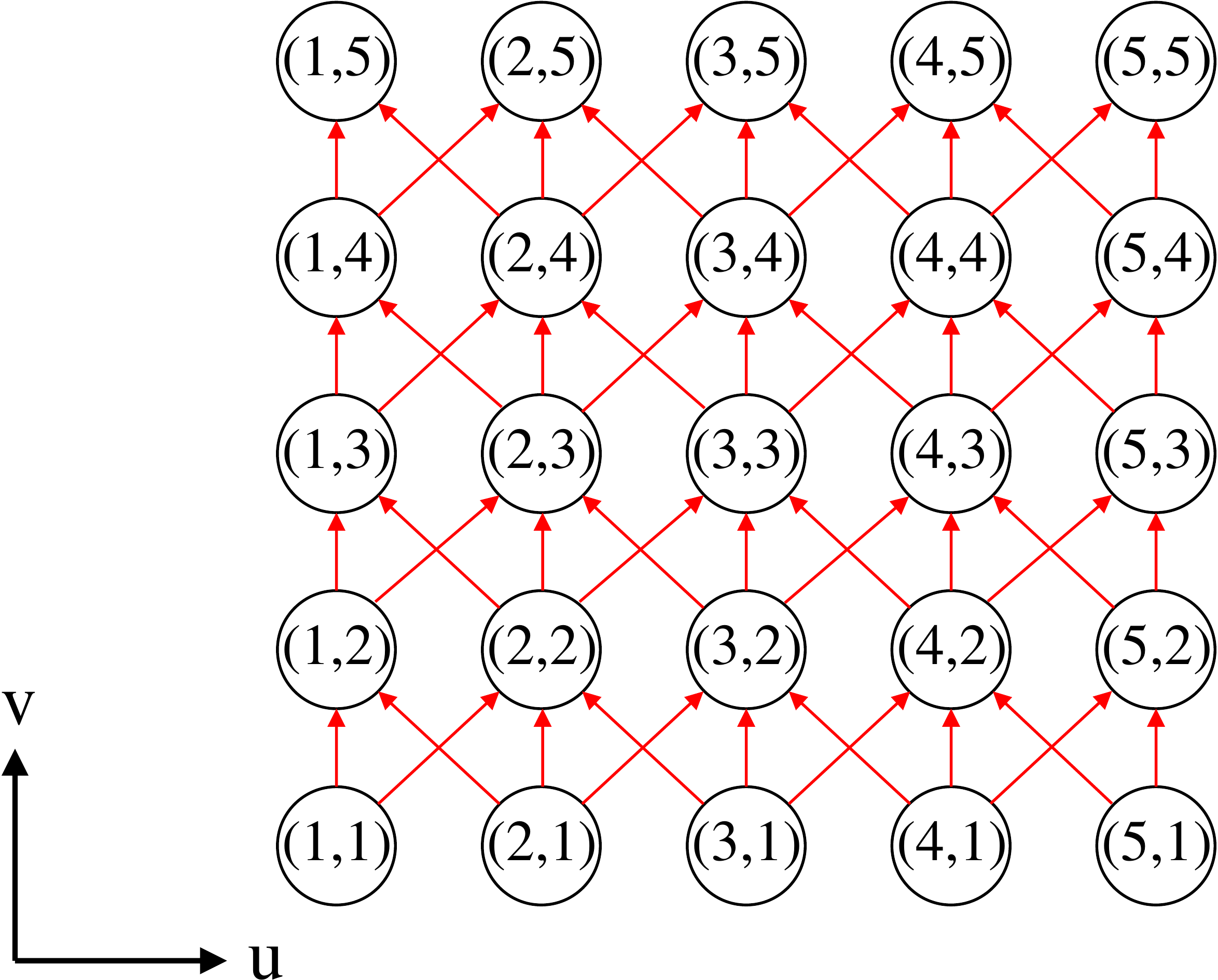}            
	\caption{The structure of a $5\times 5$ graph ($k=1$).}
	\label{fig.graph_stru}                
\end{figure}

An example of $\mathcal{G}$ is depicted in Fig. \ref{fig.graph_stru}. With assumptions and notations made, we can formulate an SSP as follows:

\textit{\textbf{Problem:}} Given a well-defined graph model $\mathcal{G}$ described above and a set of nodes $\mathcal{D}$, we need to find shortest paths from the bottom row to the destination nodes in $\mathcal{D}$. 
Please note here, \textit{shortest} is defined that a path should have the minimum cumulative cost.



SPPs have several variants \cite{magzhan2013review}: single-pair shortest path problem (SPSPP), single-source shortest path problem (SSSPP), single-destination shortest path problem (SDSPP) and all-pairs shortest path problem (APSPP). For each variation, the corresponding solutions have been proposed, e.g., the Dijkstra's algorithm and the Floyd-Warshall algorithm. The problem we aim to solve here is an SPP. Thus, we can also use the above algorithms to solve it. 
However,  several differences in the formulated problem should be considered when compared with APSPP and SDSPP:

\begin{itemize}
	\item Our problem is the degradation of APSPP. 
	We only need to find $U$ shortest paths respectively for any node on the bottom row to each node in $\mathcal{D}$.
	\item Our problem is an extension of SDSPP with $U$ destination nodes at the same time. 
\end{itemize}

In contrast, DP is more effective since this problem possesses the following properties:
\begin{itemize}
	\item Optimal substructure: the solution can be obtained by the combination of optimal solutions to its sub-problems: finding the shortest paths between two consecutive rows.
	\item Overlapping subproblems: the problem has a recursive formulation; the space of its sub-problems is small so that the solutions of the subproblems can be stored in a table to avoid recomputing.
	\label{lab.dp_property}
\end{itemize}

\subsection{Algorithm Description}
$E(u,v)$ is denoted as the minimum cumulative cost function, which indicates the cost of the shortest path when it reaches $(u,v)$. $P(u,v)$ is denoted as the position of the shortest path. We can write down an equation to recursively find the optimal $P^{*}(u,v)$ use the results to update $E(u,v)$:
\begin{align}
\begin{split}
P^{*}(u,v) 
= 
\underset{P(u,v)}{\arg\min}
\bigg\{&E\big[P(u,v)\big] \\
+&\mathcal{F}\big[P(u,v), (u,v)\big]\bigg\}\\
\text{s.t.} \ \ \ 
& P(u,v) = (u-j, v) \\
&j\in\{-k, -k+1, \ldots, k\},
\label{equ.optimal_P}
\end{split}
\end{align}
\begin{equation}
E(u,v) = E\big[P^{*}(u,v)\big] + \mathcal{F}\big[(u,v), P^{*}(u,v)\big].
\label{equ.optimal_E}
\end{equation}

With \eqref{equ.optimal_P} and \eqref{equ.optimal_E}, the SPP can be solved by utilizing DP. The detailed implementation is summarized in Algorithm \ref{alg.dp}. 
\begin{algorithm}
	\caption{The DP algorithm}
	\begin{algorithmic}[1]
		\REQUIRE $\mathcal{P}$, $\mathcal{D}$, $\mathcal{F}$
		\ENSURE $P$
		\STATE Initialization: $E(\cdot, 1)=0$, other terms are set to $+\infty$
		\STATE Initialization: $P(u,v)$
		\FOR {$v=2:V,\ u=1:U,\ j=-k:k$}
		\IF {$E(u,v) > E(u-j,v-1) + \mathcal{F}\big[(u,v), (u-j, v-1)\big]$}
		\STATE $E(u,v) = E(u-j,v-1) + \mathcal{F}\big[(u,v), (u-j, v-1)\big]$
		\STATE $P(u,v) = (u-j, v-1)$
		\ENDIF        
		\ENDFOR
	\end{algorithmic}
	\label{alg.dp}    
\end{algorithm}

\subsection{Validation}
\label{sec.graph_validation}

\begin{figure}    
	\centering    
	\includegraphics[width=0.3\textwidth]{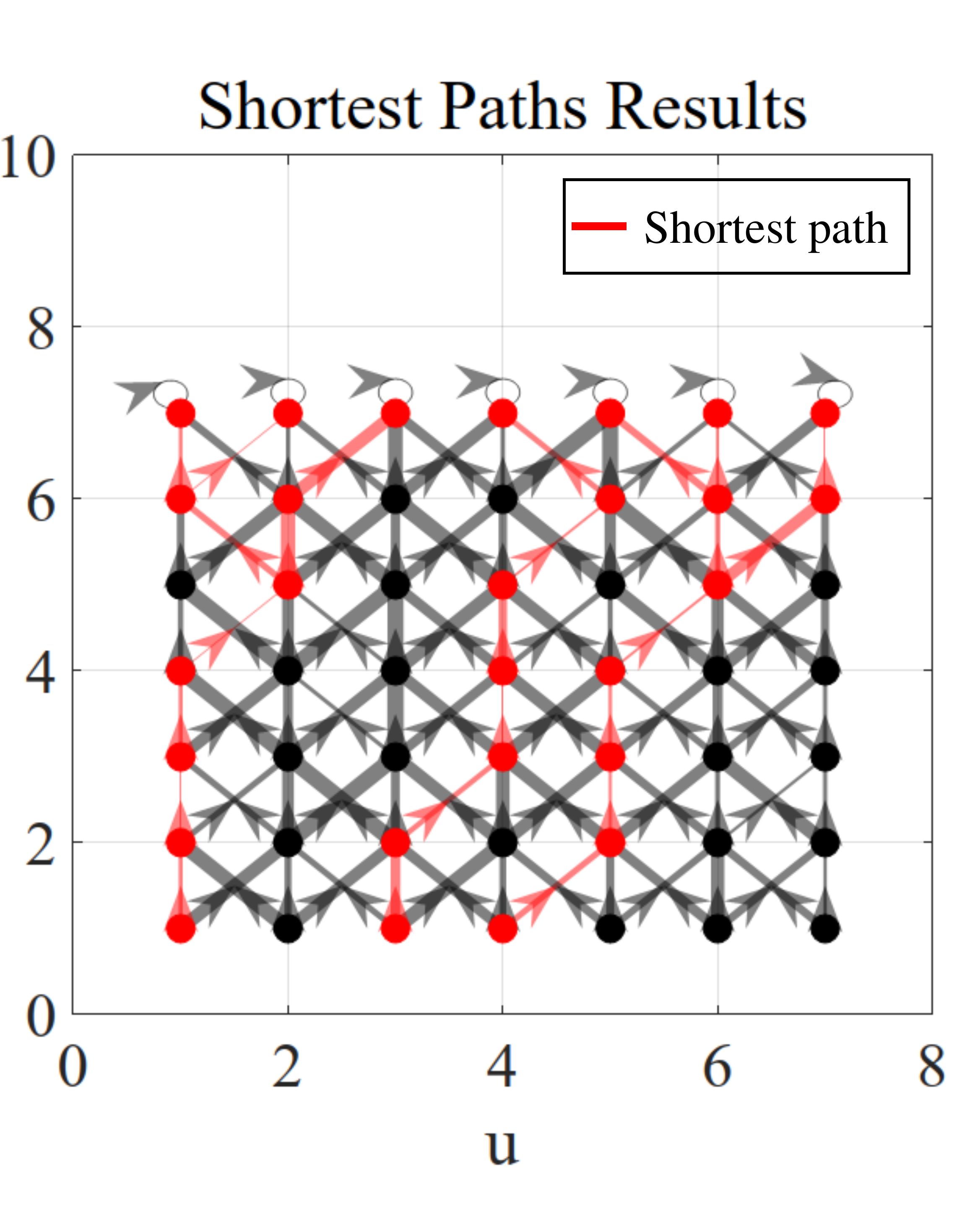}           
	\vspace{-0.3cm}          
	\caption{The results of the formulated problem in a $7\times 7$ graph are visualized. The shortest paths are searched by the proposed DP algorithm. The red edges represent the shortest paths, while the black one are the normal edges. The thickness of each edge has a linear relationship with its cost.}
	\label{fig.dp}  
\end{figure}

Fig. \ref{fig.dp} shows the shortest path results in a toy example estimated by the DP algorithm. 
To demonstrate the usage of DP in the specific graph model in terms of time and space complexity, we compare it with two general shortest path algorithms: the Dijkstra's algorithm (implemented with the Fibonacci heap) and the Floyd-Warshall algorithm. Denoting $|\mathcal{P}|$ the number of nodes, and $|\mathcal{E}|\approx k|\mathcal{P}|$ the number of edges, we can use Table \ref{tab.time_complexity} to compare the complexity of among these algorithms.
\begin{table}[h!]
	\centering
	\caption{Time complexity and space complexity comparison of the three shortest path searching algorithms.}
	\begin{tabular}{|c|c|c|}
		\hline
		Algorithm  & Time Complexity & Space Complexity\\ \hline
		Proposed DP & $O(|\mathcal{P}|)$ &  $O(|\mathcal{P}|)$ \\
		Dijkstra & $O(|\mathcal{P}|\log |\mathcal{P}|)$ & $O(|\mathcal{P}|)$ \\
		Floyd-Warshall  & $O(|\mathcal{P}||\mathcal{E}|)$ & $O(|\mathcal{P}|^{2})$ \\ \hline
	\end{tabular}
	\label{tab.time_complexity}
\end{table}

\subsection{Reasoning}
According to the time complexity and space complexity, the proposed DP algorithm is better than the others on the designed graph model. 
On the one hand, it takes advantages of the connection structure of the graph model; hence the search space is significantly reduced. 
On the other hand, DP is only applicable to problems with specific properties. Furthermore, as a typical memory intensive algorithm, DP is ineffective in a large-scale graph model.
However, considering the real applications where the graph size is usually small, meaning that our proposed algorithm is feasible in several image segmentation tasks.

\section{PRELIMINARIES}
\label{sec.preliminary}
In this section, we provide readers with mathematical preliminaries regarding VP and IPM.

\subsubsection{Vanishing Point} 
\begin{figure}
	\centering    
	\includegraphics[width=0.4\textwidth]{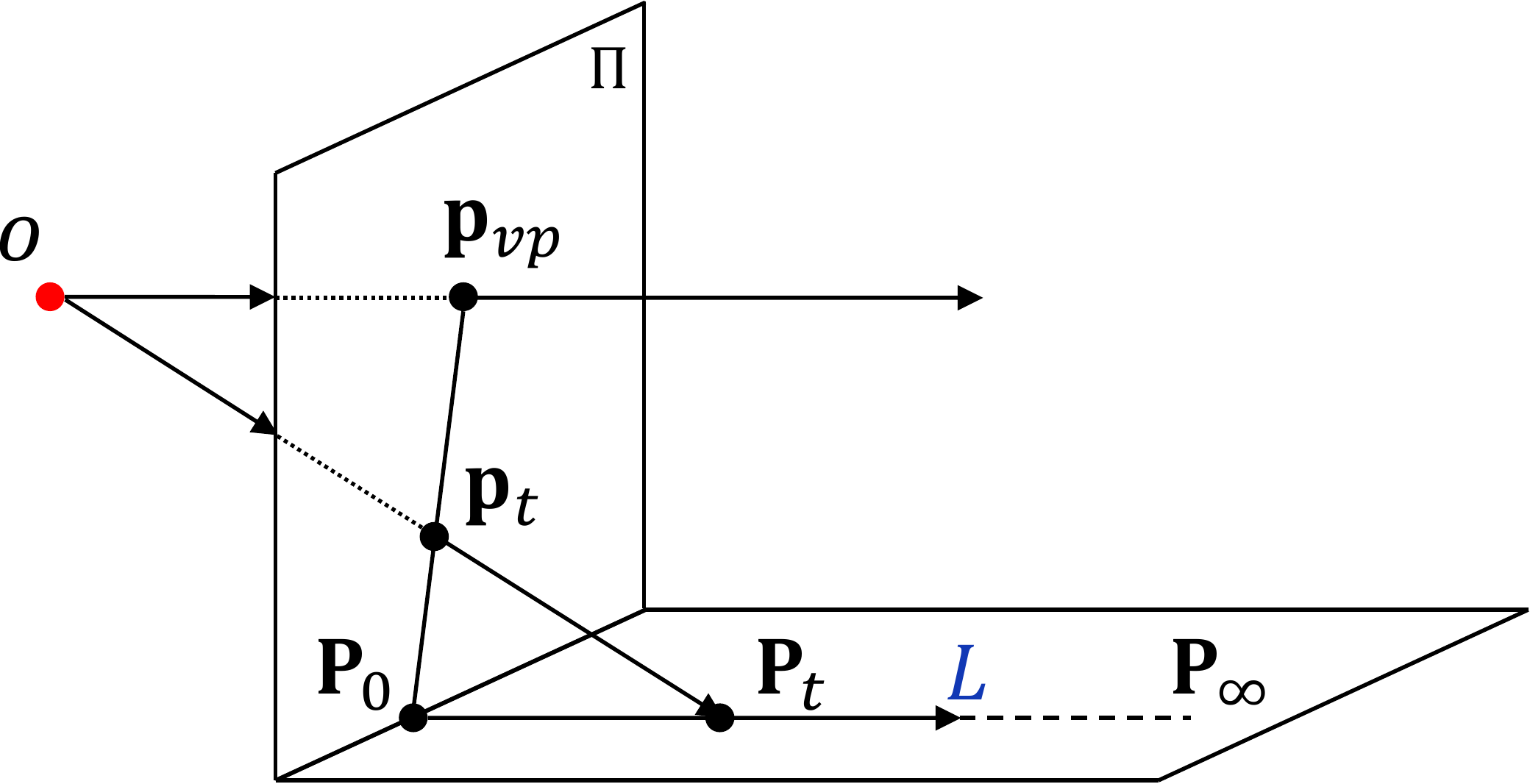}    
	\caption{The vanishing point}
	\label{vp}               
\end{figure}

The VP is produced by the perspective projection, where the projection in the image plane $\Pi$ of any pair of straight and parallel lines in $\mathbb{R}^3$ will coverage at one point. 
Fig. \ref{vp} visualizes the mentioned process. 
We denote $L$ a straight line in $\mathbb{R}^{3}$ and $\mathbf{P}_{0} = [P_X, P_Y, P_Z, 1]^{\top}$ a point in a camera cooridinate system (CCS). Therefore, the coordinates of a point $\mathbf{P}_{t}$ on $L$ are parameterized as follows:
\begin{equation}
\mathbf{P}_t = \mathbf{P}_0 + t\mathbf{n} = 
\begin{bmatrix}
P_X + tn_X \\
P_Y + tn_Y \\
P_Z + tn_Z \\
1
\end{bmatrix}
\cong
\begin{bmatrix}
\frac{P_X}{t} + n_X \\
\frac{P_Y}{t} + n_Y \\
\frac{P_Z}{t} + n_Z \\
\frac{1}{t}
\end{bmatrix},
\end{equation}
\begin{equation}
\mathbf{p}_t = \mathbf{K} \mathbf{P}_{t},
\end{equation}
where $\mathbf{n}=[n_X, n_Y, n_Z]^{\top}$ is the unit vector giving the direction of $L$, t is a scalar, $\mathbf{p}_{t}=[u,v,1]^{\top}$ is the projection of $\mathbf{P}_t$ in $\Pi$, and $\mathbf{K}$ is a camera intrinsic matrix. 

When $t\rightarrow\infty$,
\begin{equation}
\mathbf{P}_{\infty} \cong
\begin{bmatrix}
n_X \\
n_Y \\
n_Z \\
0
\end{bmatrix},
\end{equation}
\begin{equation}
\mathbf{p}_{vp} = \mathbf{p}_{\infty} = \mathbf{K} \mathbf{P}_{\infty},
\end{equation}
where $\mathbf{P}_{\infty}$ is a point of $L$ at infinity and $\mathbf{p}_{vp}=[u_{vp},v_{vp}]^{\top}$ is the VP. With similar derivation, we conclude that the projection of any parallel line to $L$ will intersect at $\mathbf{p}_{vp}$. 

\subsubsection{Inverse Perspective Mapping}
\label{sec.ipm}
\begin{figure}
	\centering    
	\includegraphics[width=0.4\textwidth]{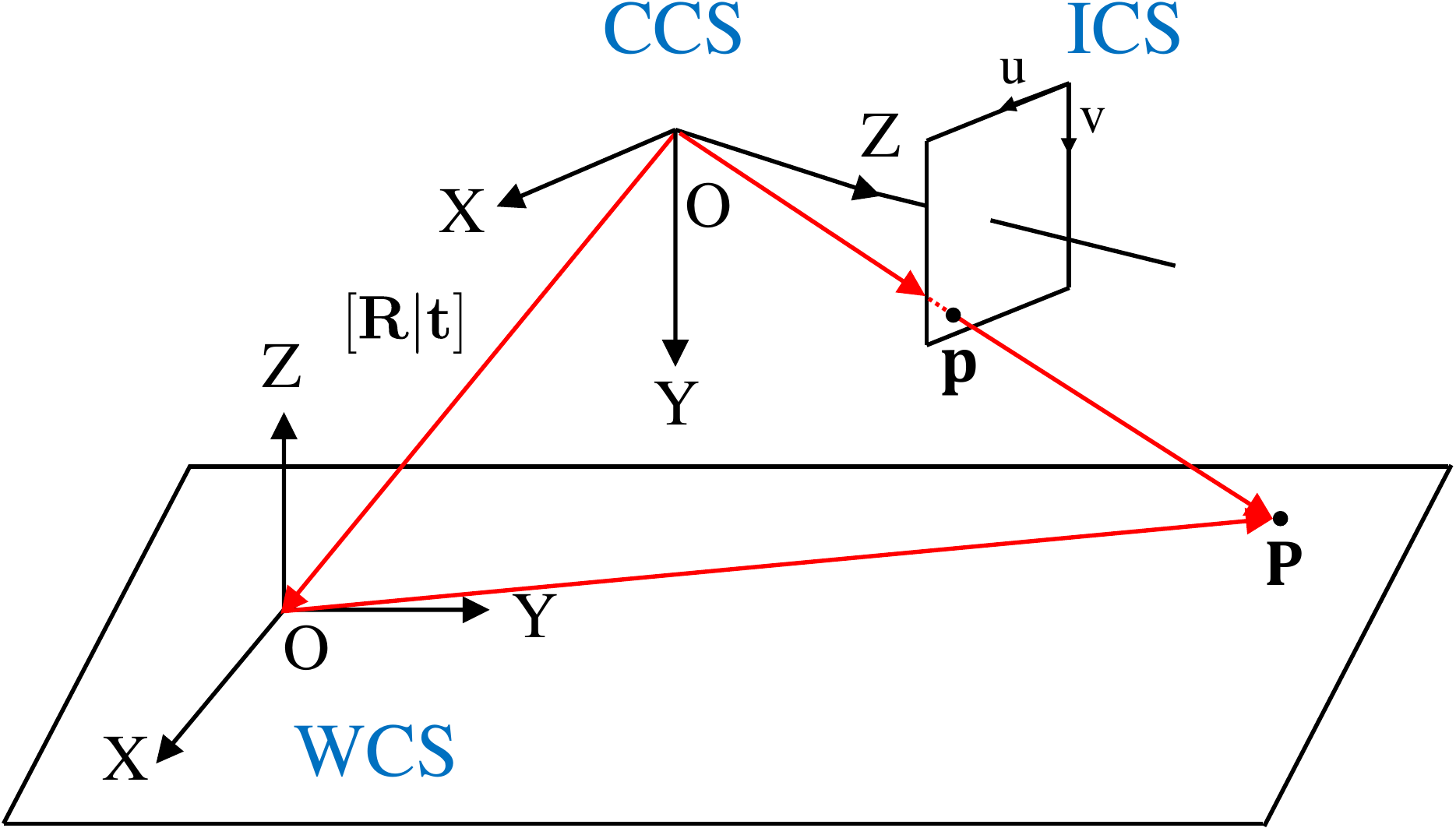}    
	\caption{The relationship between the world coordinate system (WCS), camera coordinate system (CCS) and image coordinate system (ICS).}
	\label{fig.ipm}                    
\end{figure}

IPM has been widely used for understanding road traffic signs information \cite{nieto2007stabilization}. Detailed derivation and generation of the IPM image are described in \cite{mallot1991inverse}. Fig. \ref{fig.ipm} depicts the relationship between the world coordinate system (WCS), CCS and image coordinate system (ICS). Denoting $\mathbf{P}=[X,Y,Z]^{\top}$ a point in the WCS and $\mathbf{p}=[u,v]^{\top}$ the corresponding point in the ICS. 

The transformation equation from $\mathbf{P}$ to $\mathbf{p}$ is given:
\begin{equation}
\label{equ.ipm}
\begin{bmatrix}
\mathbf{p} \\
1 
\end{bmatrix}
\sim
\mathbf{K}[\mathbf{R}| \mathbf{t}]
\begin{bmatrix}
\mathbf{P} \\
1 
\end{bmatrix},
\end{equation}
where $\mathbf{R}=\mathbf{R}_{Z}(\gamma)\mathbf{R}_{Y}(\beta)\mathbf{R}_{X}(\alpha)$.

Without loss of generality, we set $\mathbf{t} = [0, 0, -h]^{\top}$,  where $h$ is the height of the camera. We also set $\alpha=0$. If $\mathbf{p}_{vp}$ is acquired, $\beta$ and $\gamma$ can be calculated as in \cite{nieto2007stabilization}:
\begin{equation}
\beta = \arctan\left[ (\tan \alpha_v)(1-\frac{2v_{vp}}{V})\right],
\end{equation}
\begin{equation}
\gamma = \arctan\left[(\tan \alpha_u)(\frac{2u_{vp}}{U}-1)\right],
\end{equation}
where $\alpha_v$ and $\alpha_v$ are half of the vertical and horizontal angle of a camera respectively, and $V\times U$ denotes the image size.

\section{METHODOLOGY}
\label{sec.methodology}

\begin{figure*}
	\centering    
	\includegraphics[width=0.9\textwidth]{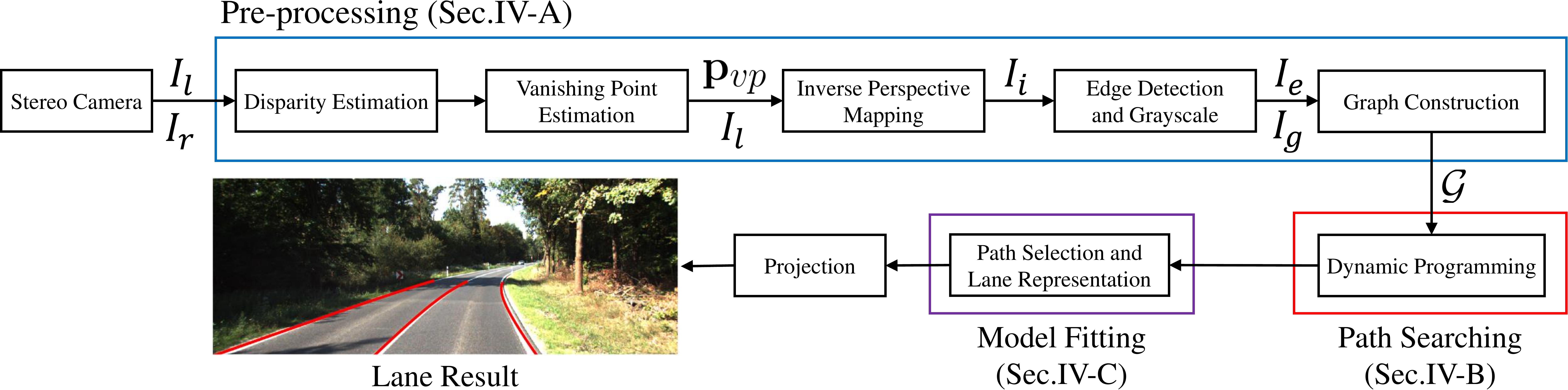}    
	\caption{Pipeline of the proposed multiple lanes detection method.}
	\label{pipeline}                    
\end{figure*}    

\begin{figure}[thpb]
	\label{fig.preprocessing}
	\centering
	\subfigure[Shortest paths in the graph]
	{\centering\includegraphics[height=0.30\textwidth]{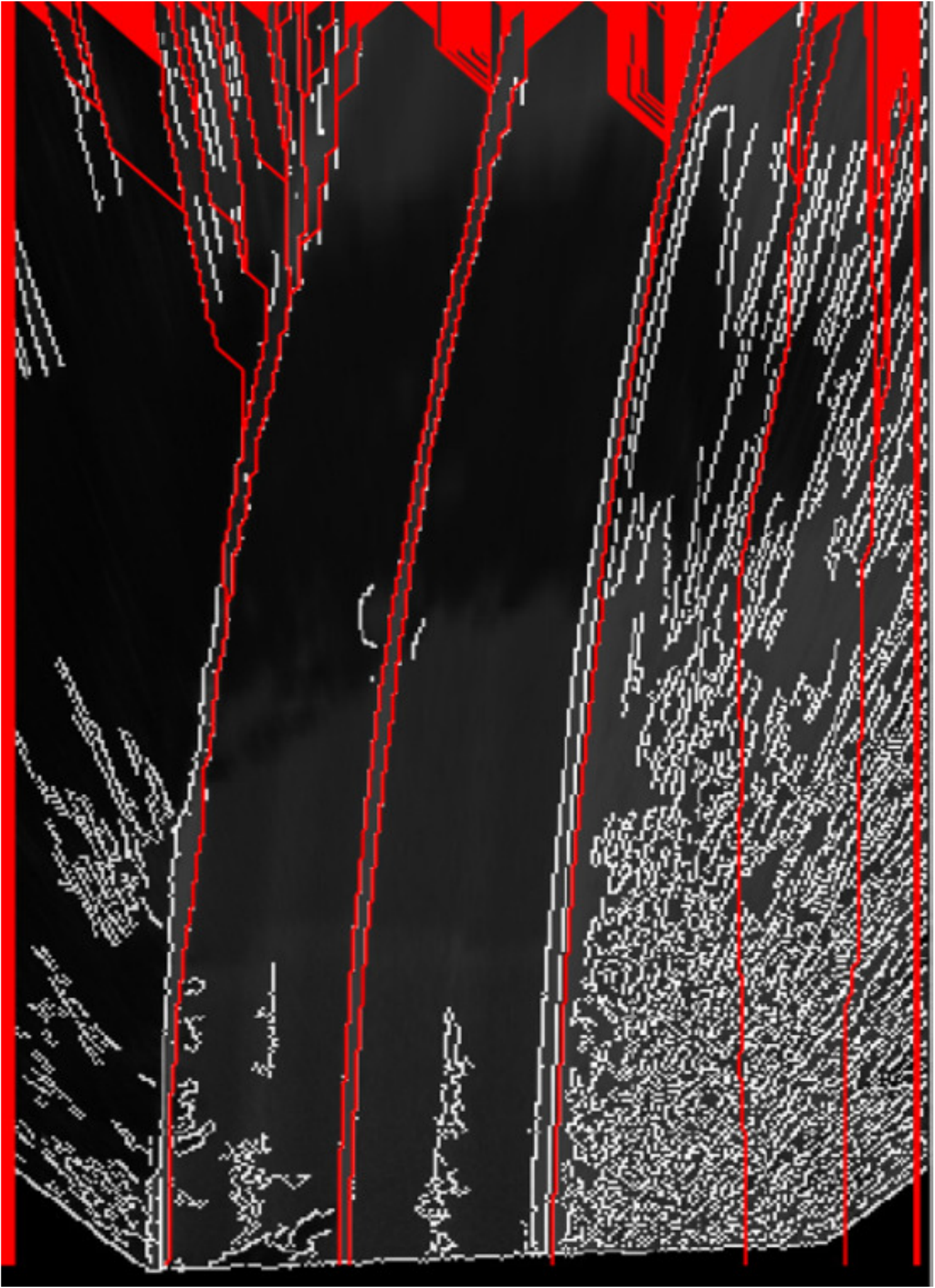}}
	\hspace{0.25cm}
	\subfigure[Lane detection result \label{fig.functionmap}]
	{\centering\includegraphics[height=0.30\textwidth]{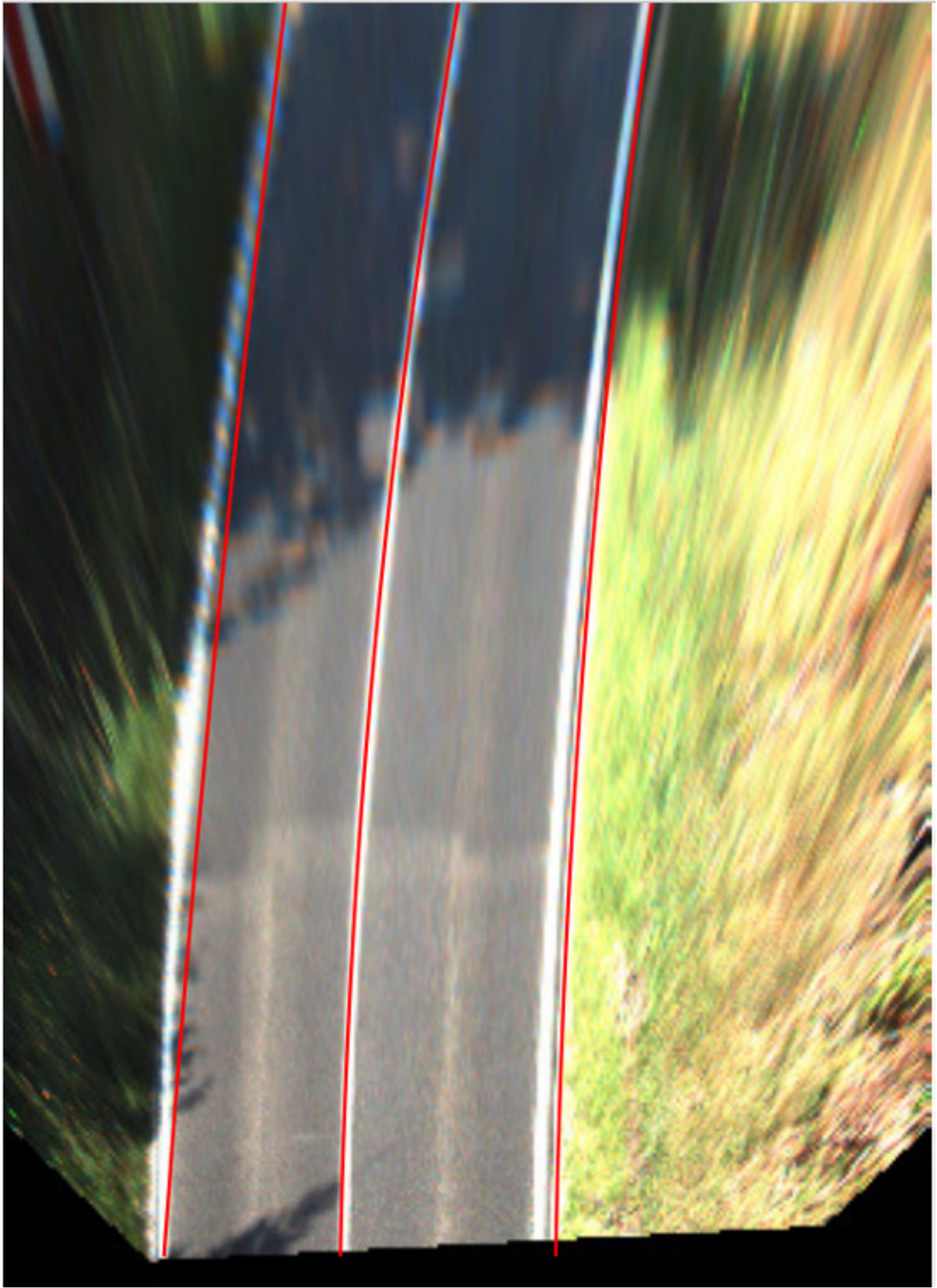}} 
	\caption{Pre-processing: (a) The Shortest paths in the graph. (b) Lane detection results.}
	\label{fig.local_cost}    
\end{figure}

In this section, we discuss how the proposed model can be applied to detect lanes on images. Fig. \ref{pipeline} shows the pipeline of the proposed methods. 
The left and right images captured by stereo cameras are denoted by $I_{l}$ and $I_{r}$ respectively, the IPM image produced from $I_{l}$ is denoted by $I_{i}$, and the VP for $I_{l}$ is denoted by $\mathbf{p}_{vp}$.

In the pre-processing module, a graph model $\mathcal{G}$ is constructed from $I_{i}$ using edge information, as depicted in Fig. \ref{fig.local_cost}. 
Then, several iterations are executed until the desired number of lanes are found in the path searching and model fitting modules. At each separate iteration, we employ DP to find the shortest path in $\mathcal{G}$ for each node in $\mathcal{D}$. 
After that, we select one path with the minimum cumulative cost. Its correctness will be checked through a parabolic model fitting. Finally, the resulting models are projected to $I_{l}$ for visualization.

\subsection{Pre-processing}
\subsubsection{Vanishing Point Estimation}
\label{sec.vpe}
VP is estimated using our previously published method \cite{Ozgunalp2017}, where two dense accumulators are optimized to extract the desirable paths with the minimum cost. The extracted paths are then interpolated into a quadratic and quadruplicate polynomial, respectively. The VP can be computed using these two polynomials. 

\subsubsection{Edge Detection}
\label{sec.edge_extraction}
The Canny detector is used to find the lane boundaries information, which includes a multi-stage process to detect a wide range of edges in images. It takes $I_{i}$ as input, and returns a binary feature function $f_{e}(u,v)$, where $u, v$ indicate the coordinates of a pixel and the pixel value is either $0$ or $1$. Specifically, we define that the pixel value of an edge is $1$.



\subsubsection{Graph Construction}
Using the color, edge, and spatial information of pixels, the original image $I_{i}$ can be easily modelled by the designed graph model $\mathcal{G}$.
For formulating lane detection as an SPP, the edge cost should be defined appropriately. 
Since lanes are typically assumed to be painted in white or yellow in high contrast, we can intuitive utilize the edge and gray value to calculate the edge cost. Before defining $\mathcal{F}(\cdot, \cdot)$, we introduce $f(u,v)$, which is computed as a weighted sum of two component feature functions as:
\begin{equation}
\begin{split}
w_{e}\cdot f_{e}(u,v) + w_{g} \cdot f_{g}(u,v)
&=
f(u,v)
\\
\text{s.t.}\ \ \ \ \ w_{e} + w_{g}
&=
1,
\label{eq.energy_function}
\end{split}
\end{equation}
where $w_{e}$ and $w_{g}$ are two scalars. The gray feature function $f_{g}(u,v)$ is defined as the normalization of $I_{g}$. Hence, the value $f(u,v)$ should be high if $(u,v)$ is located at lanes. 

With the above discussion, the value of $\mathcal{F}\left[(i,j), (s,t)\right]$ is defined as $\left[1-f(s,t)\right]$. During the process of shortest path searching, the paths will tend to pass through the lane areas.


\subsection{Path Seaching Based on DP for lane detection}
We apply the DP algorithm to a specific application: lane detection. In our implementation, we improve \eqref{equ.optimal_P} with an additional regularization term $\lambda j^{2}$:
\begin{align}
\begin{split}
P^{*}(u,v) 
= 
\underset{P(u,v)}{\arg\min}
\bigg\{&E\big[P(u,v)\big] \\
+&\mathcal{F}\big[P(u,v), (u,v)\big] + \lambda j^{2}\bigg\}\\
\text{s.t.} \ \ \ 
& P(u,v) = (u-j, v) \\
&j\in\{-k, -k+1, \ldots, k\},
\label{equ.applicationoptimal_P}
\end{split}
\end{align}
where $\lambda$ is a scalar that is defined as a sum of the square of horizontal movement. We can observe that noisy data are randomly distributed, but lane boundaries are connected continuously with strong spatial relationship. For this reason, we design $\lambda j^{2}$ to regulate the path generation in horizontal direction. On the other hand, the setting of $\lambda$ may influence the structure of the resulting paths, i.e. we can correspondingly set a large $\lambda$ to detect curved lanes.

The improved method for lane detection is detailed in Algorithm \ref{alg.dp_lane_detection}. Lines 10--14 will be explained in Section \ref{sec.path_selection}.

\subsection{Path Selection and Lane Representation}
\label{sec.path_selection}
To enhance the detection accuracy, a model fitting module is developed. In this paper, we utilize a parabolic model $f(v) = \beta_2v^{2} + \beta_1v + \beta_0$ to fit the paths, which is sufficient to represent lanes in the IPM images. 
We use $\mathbf{P}$ to indicate the path with the minimum cumulative cost, which is a $V\times 2$ matrix. The $i^{th}$ column of $\mathbf{P}$ is denoted by $[u_i, v_i]^{\top}$.
Consequently, the parameter vector $\bm{\beta} = [\beta_2, \beta_1, \beta_0]^{\top}$ can be estimated by solving a least-squares problem:
\begin{equation}
\label{eq.least_square}
\bm{\beta}^{*} 
= 
\underset{\bm{\beta}}{\arg\min}
\sum_{i=1}^{V}\Big[u_i - (\beta_{2}v_{i}^{2} + \beta_{1}v_{i}+ \beta_{0})\Big]^{2}.
\end{equation}

We employ random sample consensus (RANSAC) \cite{fischler1981random} to update the parameter vector $\bm{\beta}$ iteratively. To detemine whether a given candidate $[u_i, v_i]^{\top}$ is an inlier, the corresponding square residual $r_i = \left[u_i - f\left(v_i\right)\right]^{2}$ is computed. $n_{\text{inlier}}$ will be updated if $r_i<t_{r}$.
\begin{algorithm}
	\caption{Improved DP for Lane Detection}
	\begin{algorithmic}[1]
		\REQUIRE $\mathcal{P}$, $\mathcal{D}$, $\mathcal{F}$, $\lambda$, 
		\ENSURE $\mathcal{L}$        
		\FOR {$v=2:V,\ u=1:U,\ j=-k:k$}
		\STATE \textbf{if } $E(u,v)>E(u-j,v-1)+\mathcal{F}\big[(u,v),(u-j, v-1)\big]$
		\STATE $\textbf{\ \ \ \ \ \ \ \ \ \ \ \ \ \ \ \ \ \ \ \ \ \ \ \ \ \ \ \ \ \ } + \lambda j^{2}$ 
		\STATE \textbf{then}
		\STATE \textbf{\ \ \ } $E(u,v)=E(u-j,v-1)+\mathcal{F}\big[(u,v),(u-j, v-1)\big]$
		\STATE $\textbf{\ \ \ \ \ \ \ \ \ \ \ \ \ \ \ \ \ \ \ \ \ \ \ \ \ \ \ \ \ \ } + \lambda j^{2}$ 
		\STATE \textbf{\ \ \ }$P(u,v) = (u-j, v-1)$  
		\STATE \textbf{end if}
		\ENDFOR       
		\WHILE {\textit{maximum-value}$\big(E(\cdot, V)\big)>\epsilon$}    
		\STATE $i$ = \textit{maximum-value-index}$\big(E(\cdot, V), P(\cdot, V)\big)$
		\STATE $\mathbf{P}$ = \textit{shortest-path}\big($i$\big)
		\STATE $\bm{\beta}$ = \textit{model-fitting}\big($\mathbf{P}$\big)
		\IF {$\mathbf{P}$ \textit{has few outliers}}
		\STATE $\mathcal{L} = \mathcal{L} \cup \{\bm{\beta}\}$
		\ENDIF
		\STATE $E(i,V) = 0$
		\ENDWHILE
	\end{algorithmic}
	\label{alg.dp_lane_detection}    
\end{algorithm}

\section{EXPERIMENTS}
\label{sec.experiment}
\vspace{-0.1cm}

\begin{figure}[htp]
	{\centering\includegraphics[height=0.0485\textwidth]{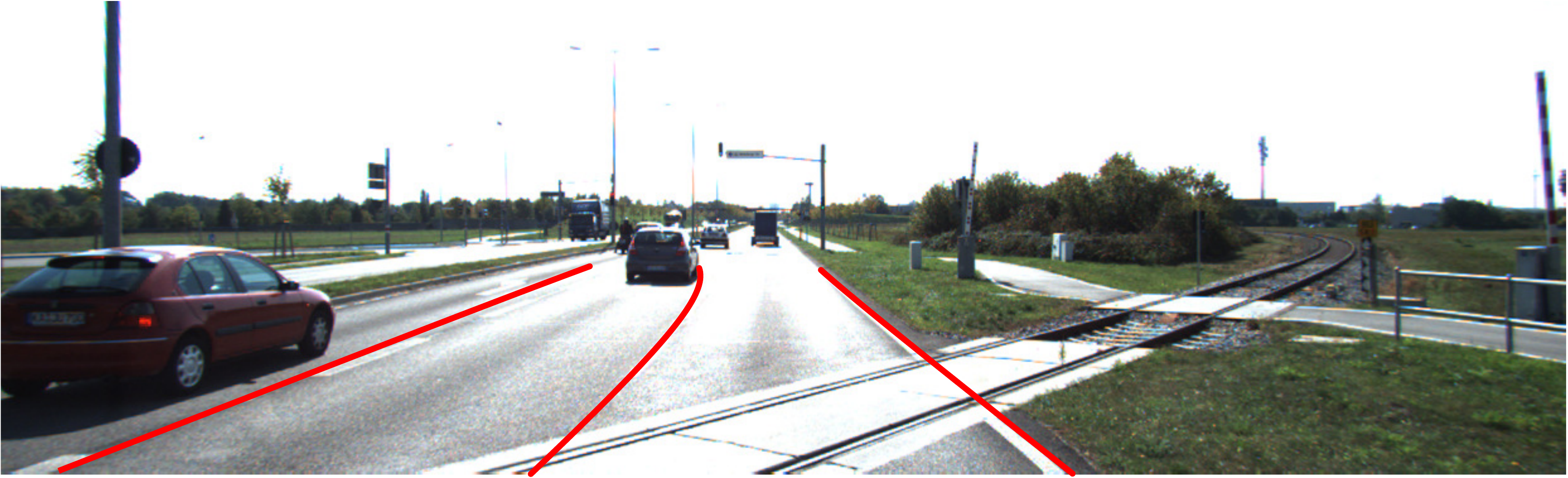}}%
	\hspace{0.05cm}%
	{\centering\includegraphics[height=0.0485\textwidth]{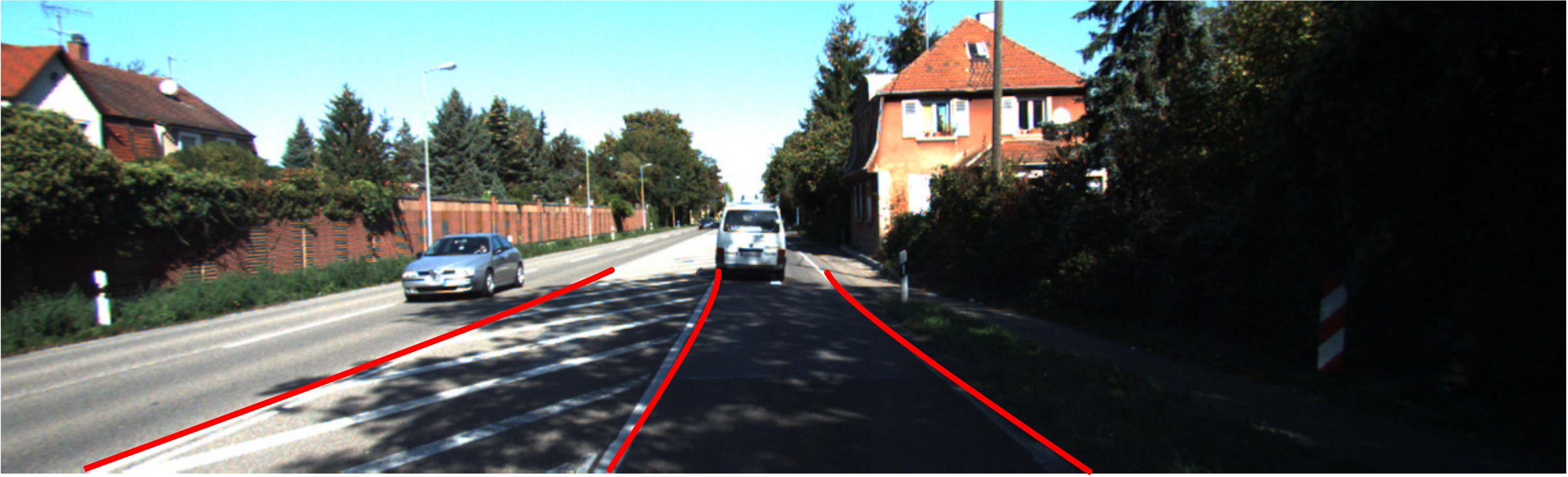}}%
	\hspace{0.05cm}%
	{\centering\includegraphics[height=0.0485\textwidth]{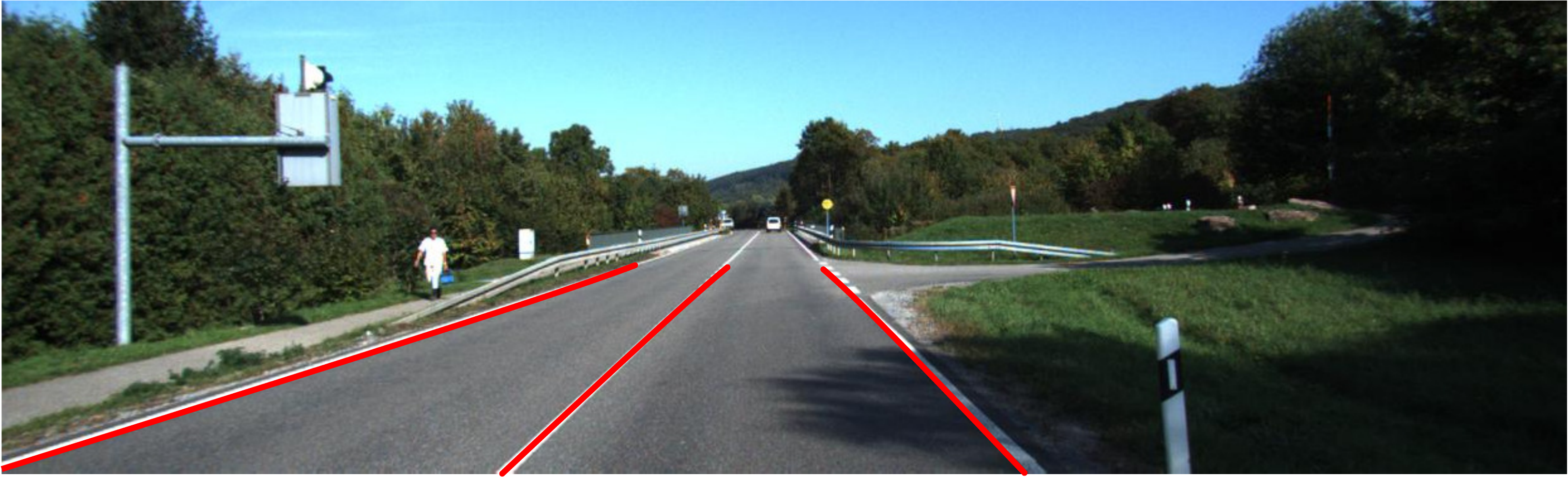}}%
	\vspace{0.05cm}
	
	{\centering\includegraphics[height=0.0485\textwidth]{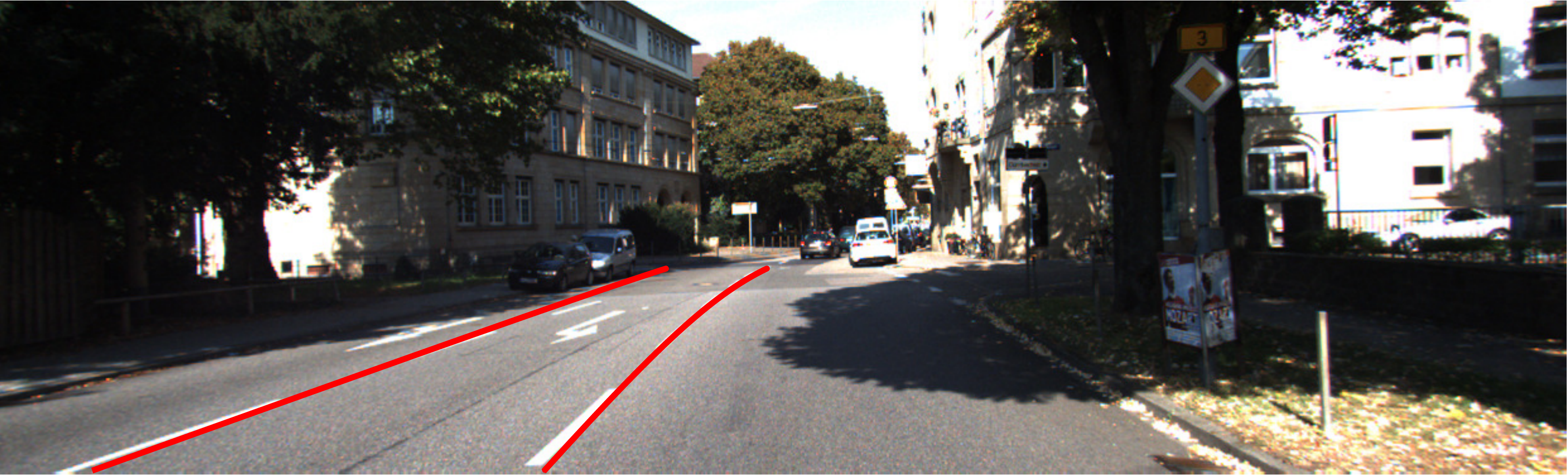}}%
	\hspace{0.05cm}%
	{\centering\includegraphics[height=0.0485\textwidth]{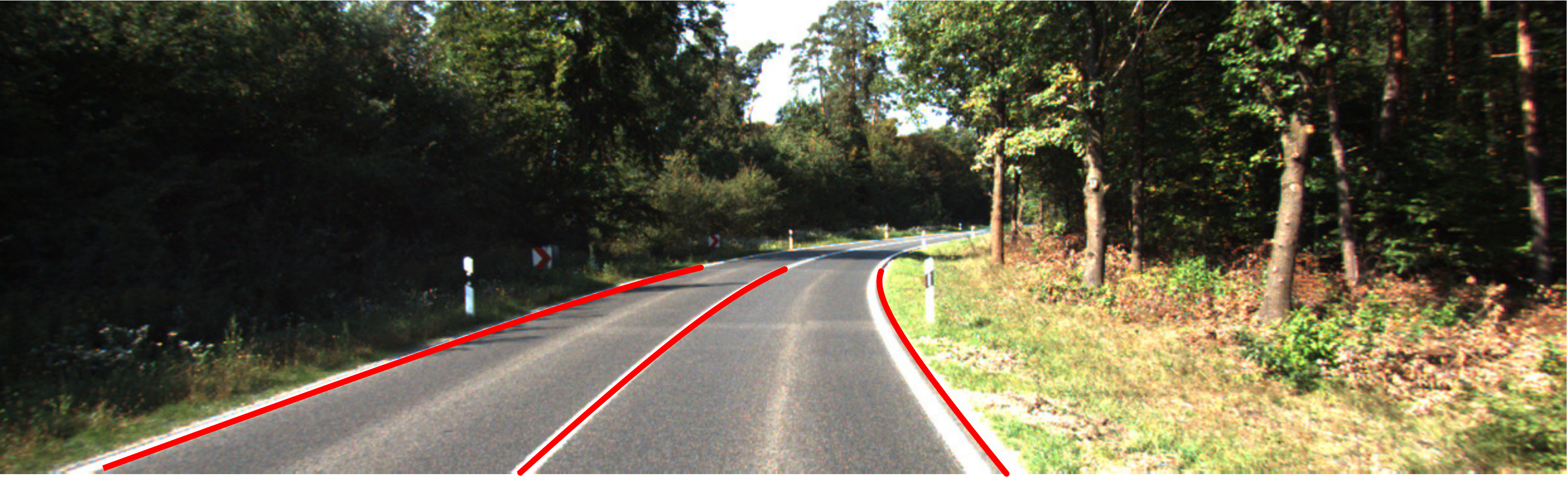}}%
	\hspace{0.05cm}%
	{\centering\includegraphics[height=0.0485\textwidth]{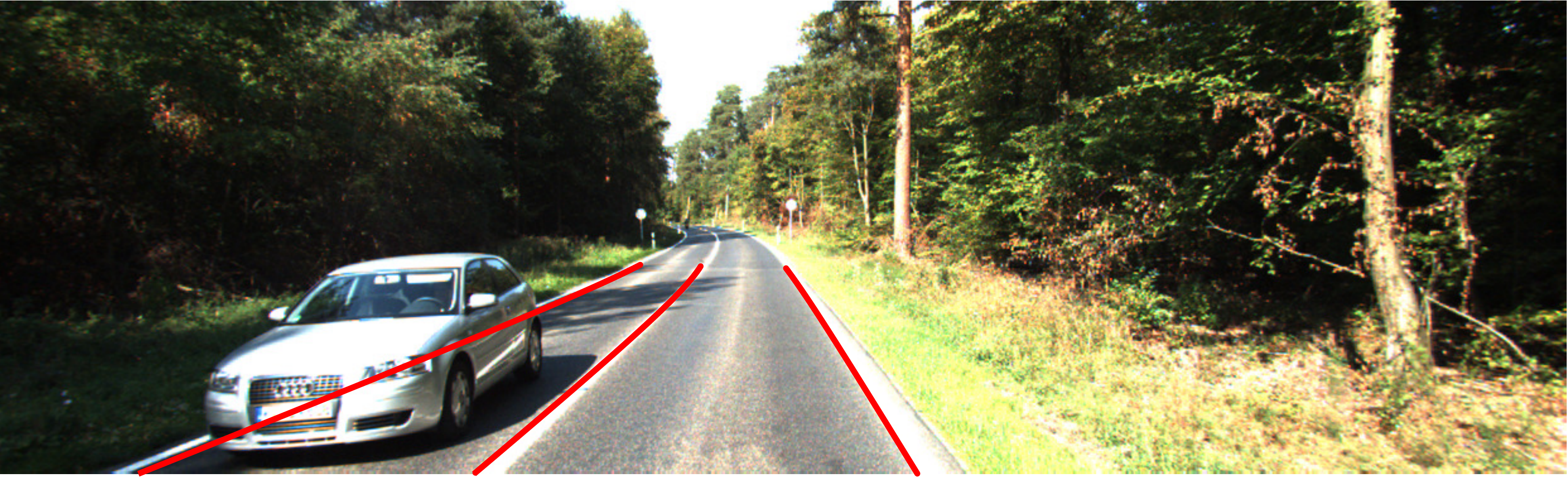}}%
	\vspace{0.05cm}
	
	{\centering\includegraphics[height=0.0488\textwidth]{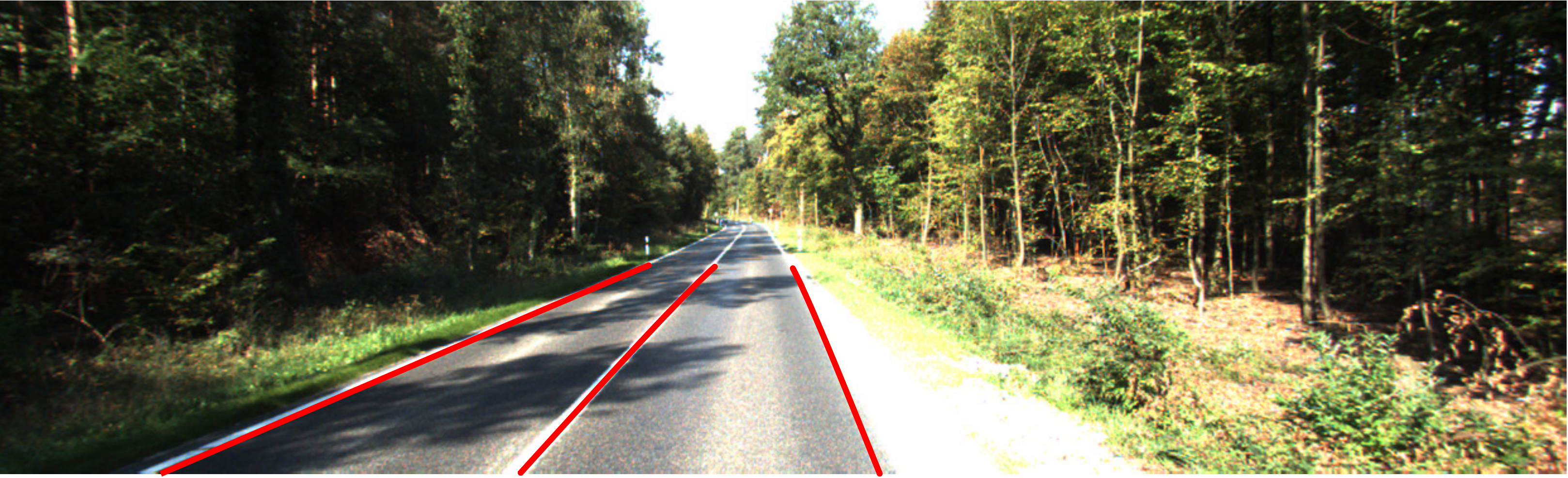}}%
	\hspace{0.05cm}%
	{\centering\includegraphics[height=0.0485\textwidth]{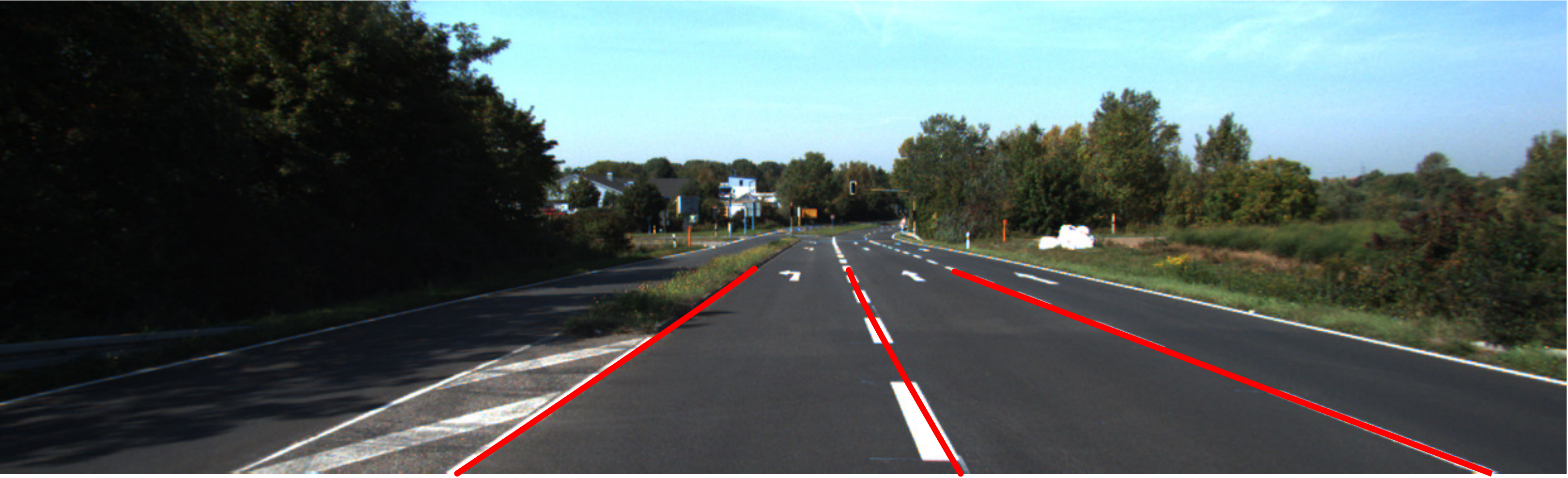}}%
	\hspace{0.05cm}%
	{\centering\includegraphics[height=0.0485\textwidth]{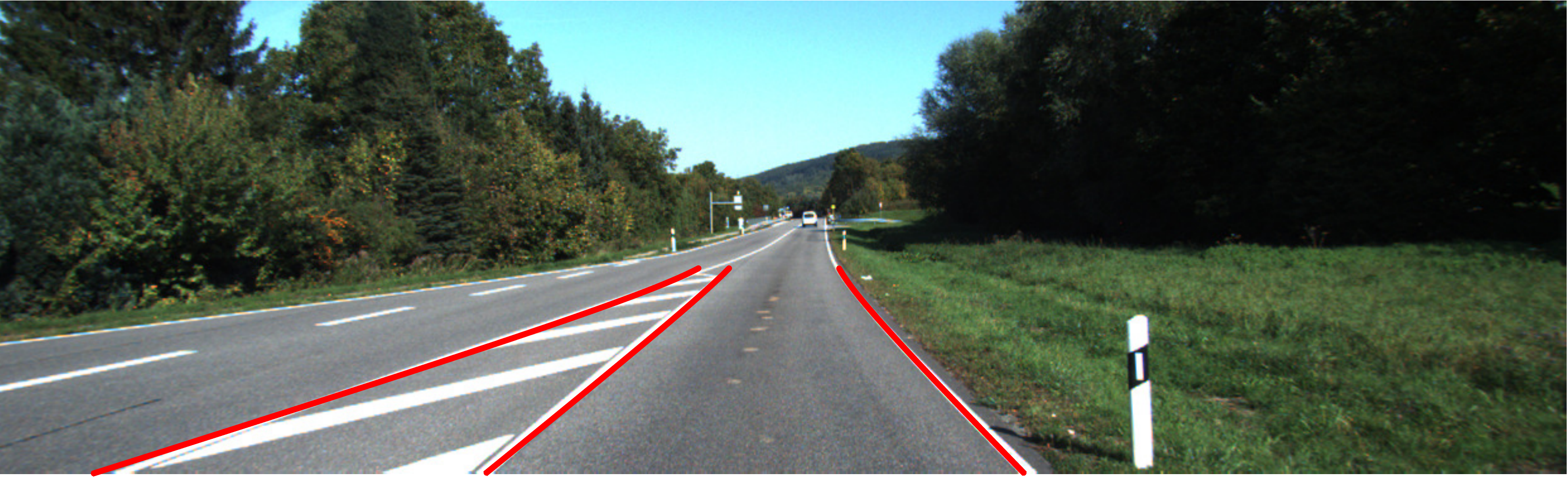}}%
	\caption{Experimental results in various scenarios, where sample detection results are demonstrated under dense shadows and high exposure. The lines in red are detected lanes.}
	\label{fig.sample_detection}        
\end{figure}

The proposed algorithm is compared with another published lane detection method \cite{Wang2012} (except for tracking step). In this method, lanes are detected based on VP estimation. All the experiments are evaluated with the KITTI dataset. 
With empirically setting the parameters $w_{e}=0.6$, $w_{g}=0.4$, $\lambda=2$, $k=3$, $t_{r}=5$, our method can obtain promising results. The IPM range covers a ground area of $30$ m and $12$ m in the Y and X directions, respectively. Lanes will be detected if they are located in this range. 

\subsection{Results}
We test the accuracy of the proposed algorithm using the KITTI dataset\cite{Geiger2013}. To evaluate its performance, we select five sequences from the KITTI dataset 
: $1573$ frames containing $3146$ lanes from the \textit{Road Category}\footnote{\url{http://www.cvlibs.net/datasets/kitti/raw_data.php?type=road}} of the KITTI dataset. 
The lane detection results of the proposed algorithm and that presented in \cite{Wang2012} are detailed in Table \ref{table.detection_ours} and Table \ref{table.detection_wang}. 
The correctness of lanes are manually checked, and only the detection of the closest two lanes are counted. In Table \ref{table.detection_ours}, the proposed algorithm presents a better precision, where $99\%$ lanes are successfully detected in the first three sequences, and the average precision in all sequences is $97.5\%$, while the precision of \cite{Wang2012} only reaches $94.5\%$. Several successful detection results are shown in Fig. \ref{fig.sample_detection}.

\begin{figure}[h!]
	\subfigure[Over-exposure\label{fig:fail_overexposure}]    
	{\centering\includegraphics[width=0.25\textwidth]{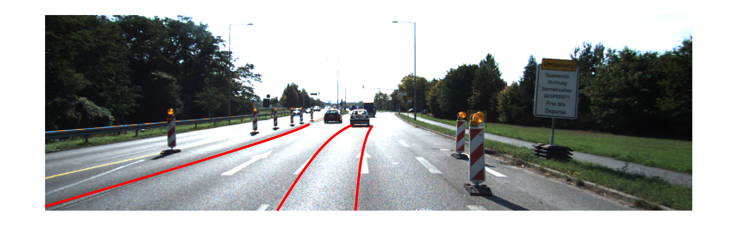}}%
	\hspace{-0.5cm}
	\subfigure[Occlusion\label{fig:fail_less}]    
	{\centering\includegraphics[width=0.25\textwidth]{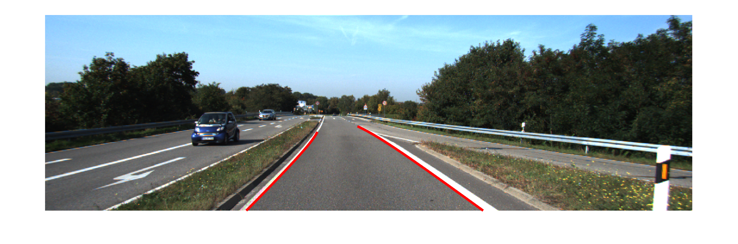}}%
	
	\subfigure[High curvature\label{fig:fail_curvature}]
	{\centering\includegraphics[width=0.25\textwidth]{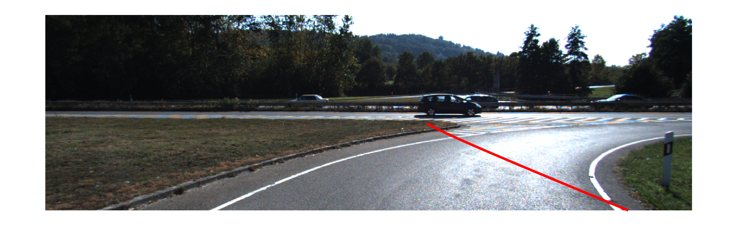}}%
	\hspace{-0.5cm}
	\subfigure[High similarity\label{fig:fail_road}]    
	{\centering\includegraphics[width=0.25\textwidth]{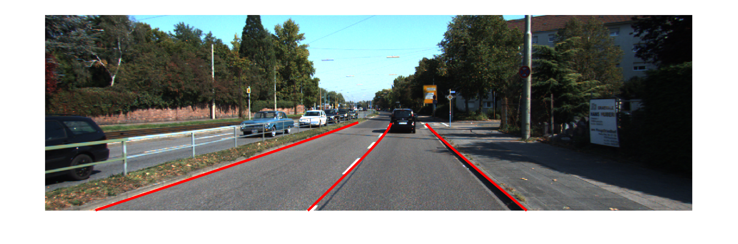}}%
	
	\subfigure[Shadow\label{fig:fail_shadow}]        
	{\centering\includegraphics[width=0.25\textwidth]{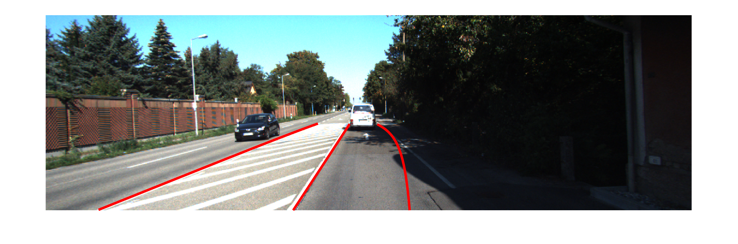}}%
	\hspace{-0.5cm}    
	\subfigure[Shadow\label{fig:fail_object}]
	{\centering\includegraphics[width=0.25\textwidth]{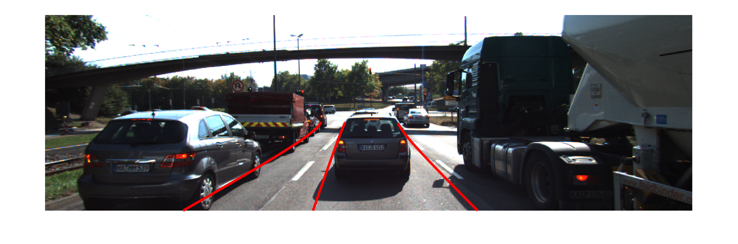}}%

	\caption{(a)--(c) illustrates the failed examples in misdetections, and (d)--(f) shows another failed examples in incorrect detections.}
	\label{fig.fail}    
\end{figure}

In our experiments, misdetections and incorrect detections of lanes induce several failed cases. The misdetections are mainly caused by camera over-exposure, partially occluded by obstacles and highly curved lanes. The detailed examples are presented in Fig. \ref{fig:fail_overexposure}--\ref{fig:fail_curvature} respectively. In Fig. \ref{fig:fail_overexposure}, due to over-exposure, lanes very obscure. In Fig. \ref{fig:fail_less}, we can see the grass partially occludes several lanes. The occlusion decreases the edge features extracted by the Canny operator. In Fig. \ref{fig:fail_curvature}, all lanes have a high curvature, making the cumulative value of $\lambda j^{2}$ large.

For the factors leading to incorrect detections, we group them into two main categories. The first category is that some objects, like road boundaries or railings, look very similar to lanes in the IPM images and sometimes fail the proposed algorithm. This case can be seen in Fig. \ref{fig:fail_road}. The other one is the shadow caused by vehicles or trees. In this case, the edges of the shadow cannot be removed cleanly, which may also fail the proposed algorithm. Fig. \ref{fig:fail_object}-\ref{fig:fail_shadow} shows two examples within this category.


\begin{table}[]
	\centering
	\caption{Detection results of the proposed algorithm.}
	\begin{tabular}{|c|c|c|c|c|}
		\hline
		Sequence & \begin{tabular}[c]{@{}c@{}}Total Lane \\ markings\end{tabular} & \begin{tabular}[c]{@{}c@{}}Incorrect\\ detection\end{tabular} & \begin{tabular}[c]{@{}c@{}}Misdetection \end{tabular} & Precision\\ \hline
		$\text{0926\_0015}$&594&4&1 & $\textbf{99.2}\%$ \\ \hline
		$\text{0926\_0027}$&376&12&2 & $\textbf{96.3}\%$ \\ \hline
		$\text{0926\_0028}$&860&1&0 & $\textbf{99.9}\%$ \\ \hline                
		$\text{0926\_0070}$&760&29&0 & $\textbf{96.2}\%$ \\ \hline        
		$\text{0930\_0016}$&556&21&7 & $\textbf{95.0}\%$ \\ \hline            
		Total&3146&67&10& $\textbf{97.5}\%$ \\ \hline            
	\end{tabular}
	\label{table.detection_ours}        
\end{table}

\begin{table}[]
	\centering
	\caption{Detection results of \cite{Wang2012}.}
	\begin{tabular}{|c|c|c|c|c|}
		\hline
		Sequence & \begin{tabular}[c]{@{}c@{}}Total Lane \\ markings\end{tabular} & \begin{tabular}[c]{@{}c@{}}Incorrect \\ detection\end{tabular} & \begin{tabular}[c]{@{}c@{}}Misdetection\end{tabular} & Precision\\ \hline
		$\text{0926\_0015}$&594&44&0 & $\textbf{92.6}\%$ \\ \hline
		$\text{0926\_0027}$&376&44&0 & $\textbf{88.3}\%$ \\ \hline
		$\text{0926\_0028}$&860&12&0 & $\textbf{98.6}\%$ \\ \hline                
		Total&1830&100&0& $\textbf{94.5}\%$ \\ \hline            
	\end{tabular}
	\label{table.detection_wang}        
\end{table}

\section{CONCLUSION AND FUTURE WORK}
\label{sec.conclusion}
In this paper, we proposed a directed graph model and a corresponding shortest path searching algorithm based on DP that can tackle a specific SPP effectively. We also presented its application to lane detection. The results showed that the designed model and the proposed DP-based algorithm with great potential in solving visual recognition problems. 

We observe that the proposed algorithm relies highly on the pre-processing results, and is thus sensitive to the changing light, occlusion and the similar appearance of other objects. But the main advantage of the proposed method is its flexibility to be a versatile solution of other visual recognition tasks.
There are several possible extensions to this work, (1) finding a way to determine the start node and destination node of a lane, (2) fusing the stereo IPM images to construct an images in a broader range from bird's eye view, (3) detecting multiple lanes at crossings and turnings, where VPs in different directions exist.

\clearpage
\balance
\bibliographystyle{IEEEtran}

\end{document}